\documentclass[conference,12pt]{IEEEtran}

\usepackage{lineno}
\usepackage{graphicx}
\usepackage{amsmath}
\usepackage{tabularx}
\usepackage{cite}
\usepackage{stfloats}
\usepackage{bm}
\usepackage{float}

\usepackage{url}

\usepackage{breakurl}
\usepackage[breaklinks]{hyperref}

\begin{document}
%

\title{Design and Implementation of a Peer-to-Peer Communication, Modular and Decentral YellowCube UUV}

\author{\IEEEauthorblockN{Zhizun Xu\IEEEauthorrefmark{1}\IEEEauthorrefmark{2}\IEEEauthorrefmark{3}, Baozhu Jia\IEEEauthorrefmark{1}\IEEEauthorrefmark{5},
and Weichao Shi\IEEEauthorrefmark{2}\IEEEauthorrefmark{6}}\\
\IEEEauthorblockA{\IEEEauthorrefmark{2}School of Engineering,
Newcastle University\\
Newcastle upon Tyne, NE1 7RU, UK\\
\IEEEauthorrefmark{1}
 School of Naval Architecture and Maritime, Guangdong Ocean University, Zhanjiang, 524000, China\\
Email: \IEEEauthorrefmark{3}Zhizun.Xu@newcastle.ac.uk,
\IEEEauthorrefmark{6}Weichao.Shi@newcastle.ac.uk}
}
\maketitle

\begin{abstract}
The underwater Unmanned Vehicles(UUVs) are pivot tools for offshore engineering and oceanographic research. Most existing UUVs do not facilitate easy integration of new or upgraded sensors. A solution to this problem is to have a modular UUV system with changeable payload sections capable of carrying different sensor to suite different missions. The design and implementation of a modular and decentral UUV named YellowCube is presented in the paper. Instead a centralised software architecture which is adopted by the other modular underwater vehicles designs, a Peer-To-Peer(P2P) communication mechanism is implemented among the UUV's modules. The experiments in the laboratory and sea trials have been executed to verify the performances of the UUV. 
\end{abstract}


\IEEEpeerreviewmaketitle

\section{Introduction}

Over the past few decades, the Unmanned Underwater Vehicles(UUVs) have become the essential tools in the offshore engineering and the ocean research. Their tasks ranges from the offshore engineering, oceanographic research, salvage and rescue to the military monitoring. As missions are becoming more and more diverse, the vehicles are supposed to be equipped with various scientific payloads or specific-designed manipulators to meet particular needs from different scenarios. Meanwhile, with advancement in technology, sensors and actuators with higher performances will be requested for conducting more efficient operations. For existing underwater vehicles, it is not easy to integrate the latest devices or adapt to specified actuators without making significant changes to the mechanical and software system\cite{sangekar2008hardware}.

In this context, modular structures for marine systems have significant advantages. That would not only reduces the modification time consumption, but also simplifies the adaption of the UUVs for various underwater applications\cite{mangayarkarasi2024modular}. Modules for different functions or measurements are constructed separately and interchanged for different missions as per requirements. In case of failures, individual modules can be easily replaced and debugged, which significantly reduces down time. 

To achieve the goal, a Peer-to-Peer(P2P) communication, modular and decentral underwater vehicle YellowCube has been developed. Unlike existing platform, the modularisation designs are not only on hardware structure, but also on software development. The entire vehicle has been separated into several modules depending on distinct functions. In the hardware level, various modules are separated into various water-tight enclosures physically to execute the particular function. All modules have their own micro onboard computers and the generic communication and power interfaces. At software development,  an inter-communication network across modules is developed on the top of the ROS framework. That means each module owns its ROS node, which can independently publish or subscribe messages in the communication network. In the case, the P2P communication mechanism has been built up among the modules, so that any module can talk or listen to an arbitrary module. The vehicle operators also are able to communicate with any module of the vehicle directly. The advantages for such design has been summarised as follows, 

\begin{itemize}
    \item The failure of one module will not affect other modules;
    \item The module can be integrated or decoupled quickly;
    \item The software of the modules can be maintained and updated individually. 
\end{itemize}

In this paper, the mechanical and software layouts of the YellowCube UUV has been presented. The essential modules in the vehicle are demonstrated in detail. At the end, the performances of the vehicle in laboratory experiments and the sea trial are discussed.

\section{Literature Review}


There are some pioneering works about UUVs modularisation in the hardware level. Sangekar, Chitre, and Koay have developed a modular autonomous underwater vehicle \cite{sangekar2008hardware}. Each section of the vehicle has a standard mechanical interface for inter-module connection. A single centralized onboard computer runs the configuration server in the Ethernet-based communication network between modules. 
In contrast to the physical connections among modules, Tolstonogov uses the Bluetooth 5.0 Mesh technology for wireless inter-modular communication and wireless power transfer system, to develop a new modular underwater vehicle, that is the complete separation of an underwater vehicle on different self-sustaining modules with its control and power system\cite{tolstonogov2020modular}. It indicates a module is connected with nearby ones only by mechanical fastenings without any electrical or communication connection. Apart from modularising the whole vehicle, a total independent propulsion module with power, communication, micro-computer, sensors(GNSS, IMU), and directional thrust within a compact, watertight container, is proposed as well\cite{odetti2024minion}.

Meanwhile, many researchers have reported excellent studies about modular software designs for underwater vehicles. Although there few software developed based on other frameworks\cite{ramey2018modular}\cite{meinecke2011hybrid}, the MOOS(Mission Orientated Operating Suit)\cite{newman2008moos} and ROS(Robotic Operation System)\cite{macenski2022robot}\cite{quigley2009ros} are two essential tools which are widely adopted by developers. Both of them are sort of open-source frameworks that helps developers create robotic software by providing tools, libraries, and conventions, in order to simplify the development of robotic applications by providing a standardized framework for connecting and controlling different components. Though both of them have similar publish-subscribe massaging communication mechanisms, the ROS is a Peer-to-Peer network and the MOOS relies on the like-mailbox data transferring mechanism\cite{demarco2011implementation}. That makes the communication of the ROS more efficient. 

However, the MOOS has historically been popular within the underwater robotics community. It provides libraries containing a platform-independent, inter-process communication API, sensor management such as DVL, LBL and SAS sonar, and automation tools such as the PID controller and the navigation system. There are many applications of marine vehicles developed based on the MOOS \cite{turrisi2024decentralized}\cite{gershfeld2023adaptive}\cite{paine2024model}. Meanwhile, because of its more advanced inter-communication mechanism and large community support, the ROS becomes the essential tool for developers who are developing the modularised software architecture for robotics. 

In order to benefit from the merits of the MOOS and the ROS, a ROS/MOOS bridge is proposed in \cite{demarco2011implementation}, which can link the MOOS database to the ROS core. In the paper, the Yellowfin AUV is introduced, where the low-level control and decision-making process is managed by the MOOS, while the image processing is charged by the ROS. As the ROS is being accepted widely by the academic grounds and research, more and more researchers exclusively utilise the ROS to develop the modular software system for marine vehicles. In \cite{naglak2018backseat}, authors have written a ROS driver for SandShark, which interacts with the critical sensors and low-level actuators, and used the computer vision packages in ROS libraries directly to percept the environment. In the subjuGator Project\cite{bjellossubjugator}, the ROS is employed to build whole software stack of the AUV. Furthermore, a customised ROS nodes have been used to process the data from the imaging sonar and the side-scan sonar\cite{aaltonen2020implementation}.

As the ROS2 has released, which is expected to improve the performances of real-time applications dramatically, Flores and Bachmayer give an interesting conception for a modular and central software system of AUVs, which is built on top of ROS2 framework. The micro-ROS running on the microcontroller(STM32F405) manages low-level functions, while the main ROS2 running on the Nvidia Jeston Nano is responsible for heavy duties ranging from providing external communication via the Ethernet to processing sensors(DVL, IMU) data. However, only does the sophisticated concept present, instead of demonstrating a concrete vehicle.


In the above works, the modularisations in software and in mechanical structure are realised independently. Oppositely, the mechanical and software modularisations on the YellowCube are consistent. Instead a centralised software architecture, the P2P communication mechanism is implemented among the YellowCube's modules. Such configuration makes the YellowCube UUV be with more advances in the updating and maintenance of the software system, quick integration of devices, and the fault tolerance.   

\section{YellowCube UUV Overview}

\begin{figure}[H]
    \centering
    \includegraphics[scale=0.03]{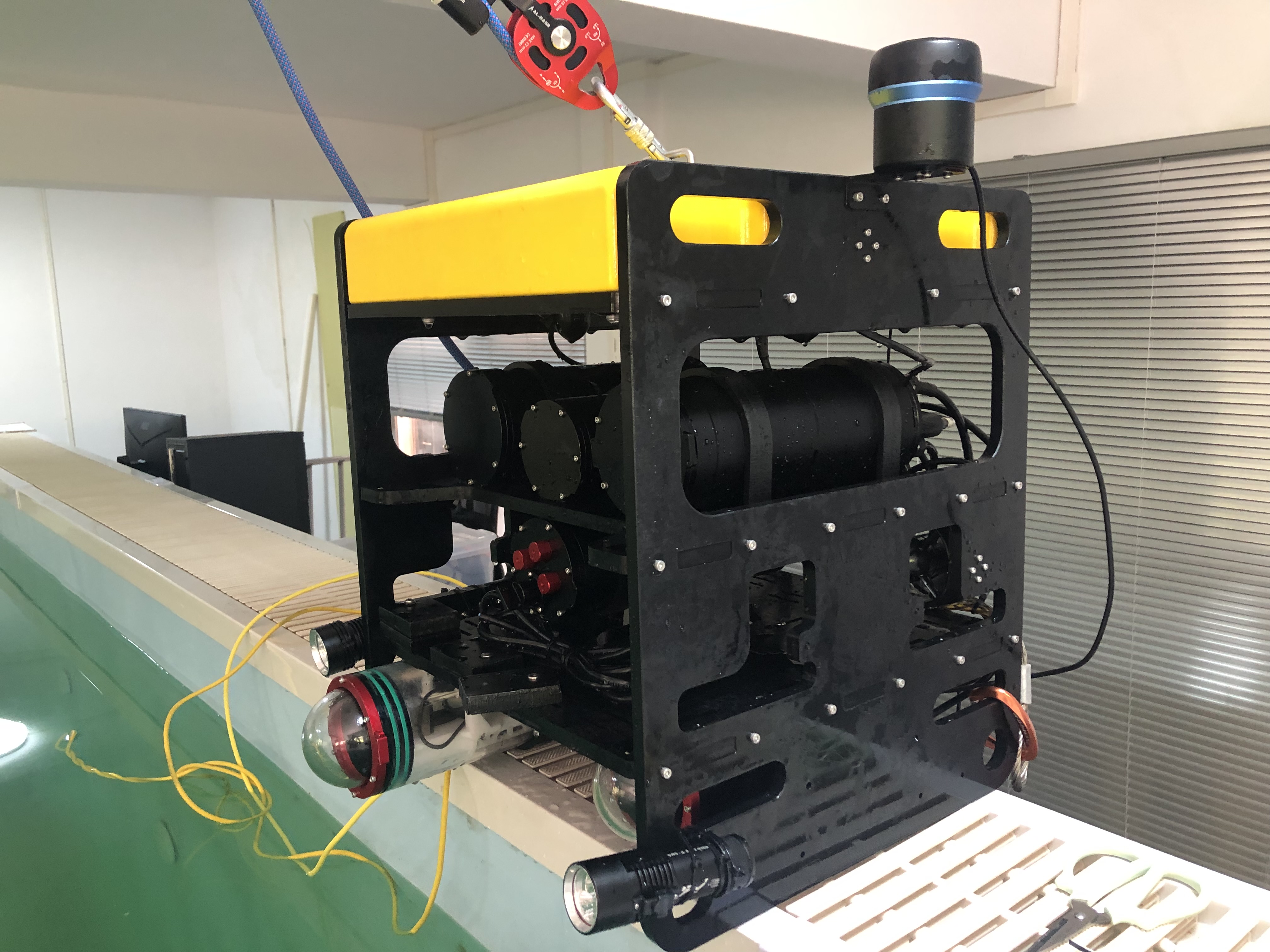}
     \includegraphics[scale=0.2]{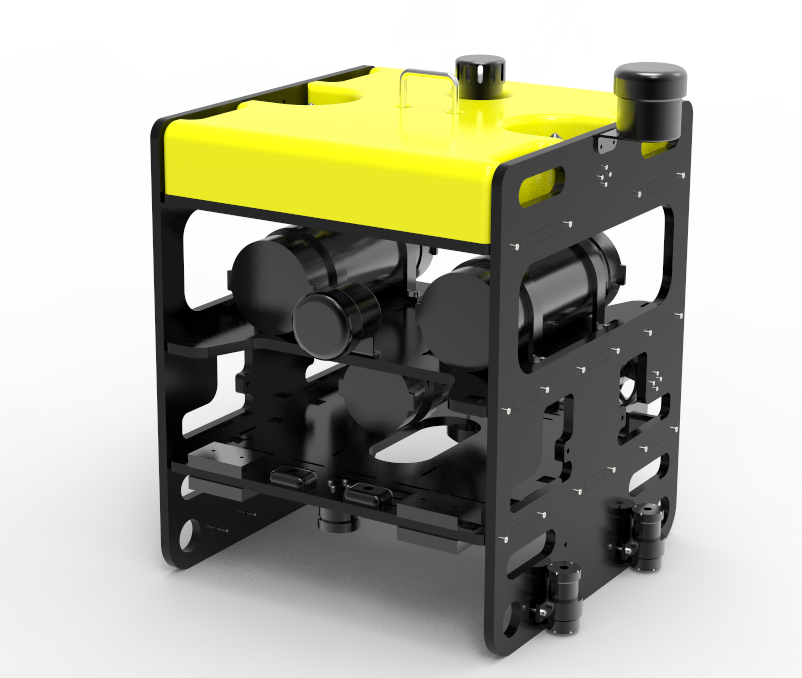}
    \caption{YellowCube UUV(Right: the real vehicle. Left: 3D rendering of the vehicle)}
    \label{fig:myROV_image}
\end{figure}

As shown in Fig.\ref{fig:myROV_image}, YellowCube is an open-frame UUV. The three core modules are fasten on the UUV frame, which are sensing module, the low-level control module, and navigation module. Except core modules, the customised scientific payload can be integrated flexibly. The frame of YellowCube UUV frame is made of the black High Density Polyethylene(HDPE). It is a polyethylene thermoplastic made from petroleum. The buoyancy material is epoxy /hollow glass microsphere syntactic foam, which provides elevated compressive strength, ultra-low density and low moisture absorption coefficient\cite{christ2013rov}. With the configuration, the vehicle can dive into 500 meters depth. The hardware information of the UUV is presented in Tab.\ref{tab:config_UUV}.

\begin{table}[H]
    \centering
    \caption{Configuration of YellowCube UUV}
    \begin{tabularx}{0.9\linewidth}{c||X}
    \hline
       Dimensions & 0.45m x 0.45m x 0.55m  \\
       Weight  & 50Kg\\
       Thrusts & BlueRobotics T200 x 4 \\
       Sensors &  IMU, DVL, USBL, Pressure Sensor, Altimeter, Camera \\
       Communication &  Power Line Carrier \& Optical Fiber \\
       Depth & 500 meter \\
       Power & Li-Ion Battery 24.4V 28AH \\
    \hline
    \end{tabularx}
    \label{tab:config_UUV}
\end{table}

\begin{figure}[H]
    \centering
    \includegraphics[width=0.9\linewidth]{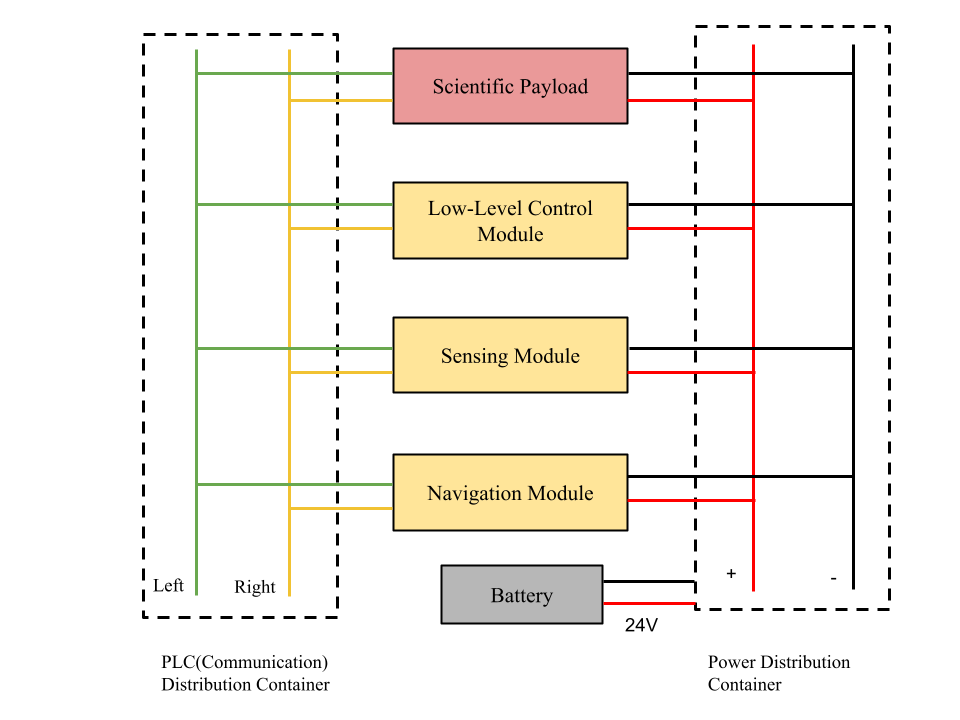}
    \caption{Mechanical Architecture}
    \label{fig:mechanical_architecture}
\end{figure}

\begin{figure}[H]
    \centering
    \includegraphics[width=0.8\linewidth]{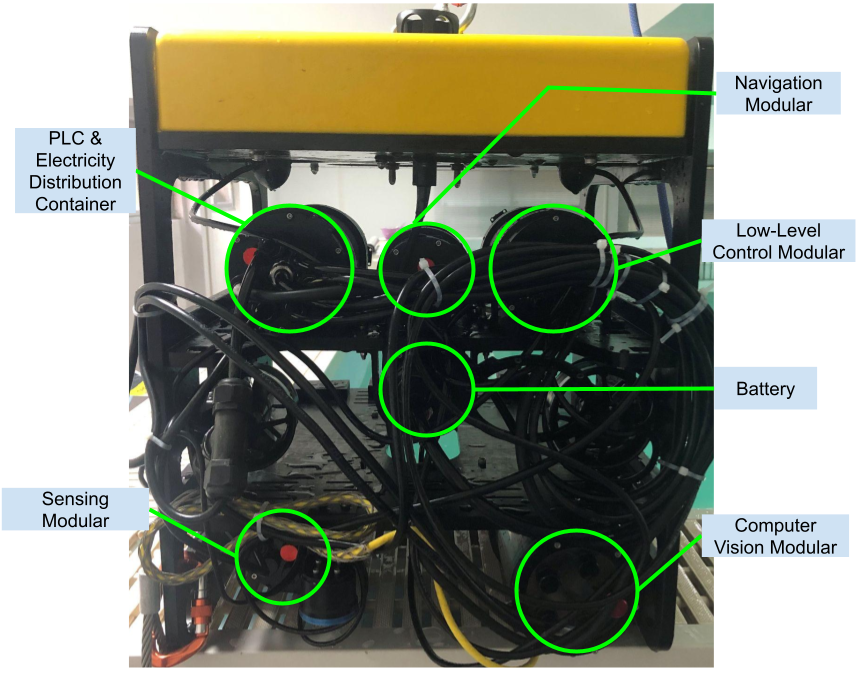}
    \caption{The Locations of Modules on The Vehicle}
    \label{fig:modulesOnVehicle}
\end{figure}

The modules are powered by a Li-Ion battery, specifically, the battery connects to the electric distribution container. The latter will be responsible for power supply for each cable-connected modules. The communication between modules is implemented by the using Power Line Communication(PLC) technology, which transmits data over existing electrical power lines. Though communication can be realised by the power cable as well, the extra distribution container for PLC cables is setup in sake of high-performance communication. As preceding mentioned, the modules are fasten on the frame. In implementation, each module connects to the power and communication containers in parallel via the generic communication and electric supply interfaces, as shown in Fig.\ref{fig:mechanical_architecture} and Fig\ref{fig:modulesOnVehicle}.  

\newpage

The software system is developed on the top of the ROS1 framework. Although the latest ROS2 has outperformed in real time applications, the better compatibility of the ROS1 makes it a ideal tool in the project which is not sensitive in the timing of reactions.
In order to set up a P2P network, each module owns a micro-computer(Raspberry Pi) running a ROS node for publishing or subscribing messages. Hence, the module is represented by its ROS node in the software level. Such configure allows every module has an ability to build up the the network and talk or listen to other modules. The vehicle operator can access the modules directly via their ROS nodes. 

\begin{figure}[H]
    \centering
    \includegraphics[width=0.8\linewidth]{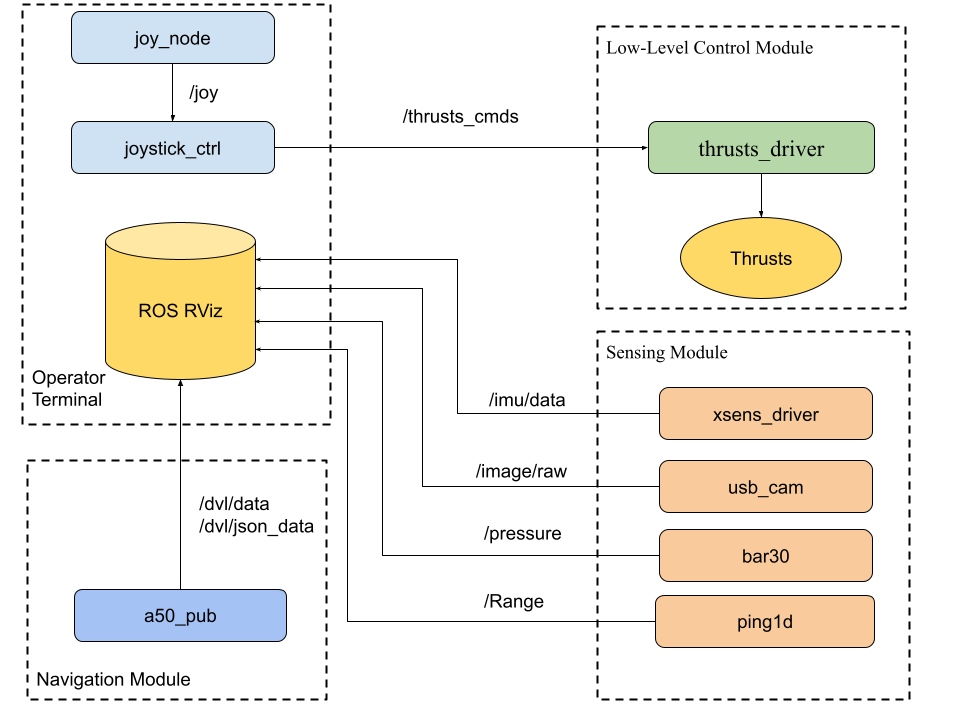}
    \caption{Software Architecture}
    \label{fig:software_architecture}
\end{figure}

\begin{figure}[H]
    \centering
    \includegraphics[width=0.6\linewidth]{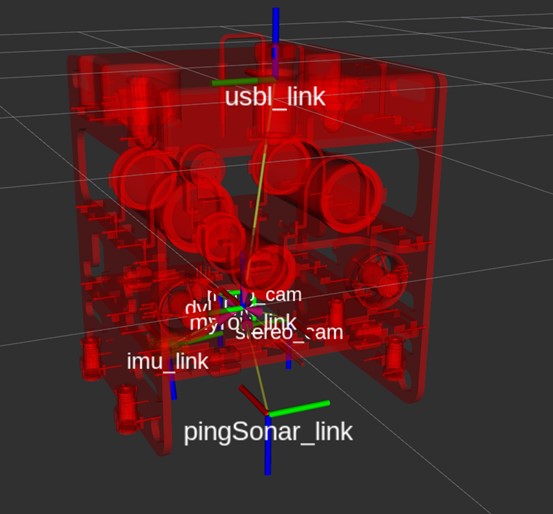}
    \caption{The States of YellowCube UUV in RViz}
    \label{fig:vehicleInRViz}
\end{figure}

The ROS nodes graph is presented in Fig.\ref{fig:software_architecture}. The operator terminal is the ROS environment running on the control platform for UUV operators, which often locates on the shore or a support ship. The operators terminal runs ROS RViz listening from multiple modules, and joystick nodes(joy\_node and joystick\_ctrl), sending the control commands to the low-level control module via a /thruster\_cmds message. In the level-low control module, a ROS node referring to as thrusts\_driver will be running and listening to the message from joystick\_ctrl which converts the joys commands to thrust commands. The module is used to control four thrusts. Similar multiple ROS nodes(xsens\_driver, usb\_cam, bar30, and ping1d) running in sensing module and one single node(a50\_pub) running in Navigation module will talk to ROS Rviz in the operator terminal. The Rviz is a 3D visualisation tool that allows users to observe and interact with the robot's environment and its internal data in a graphical interface\cite{rviz360aadithya}, as presented in Fig.\ref{fig:vehicleInRViz}.  

\section{Core Modules}

In the section, the three core modules, low-level control module, sensing module, and navigation module are presented in detail.

\subsection{Low-Level Control Module}

\begin{figure}[H]
    \centering
    \includegraphics[width = 0.8\linewidth]{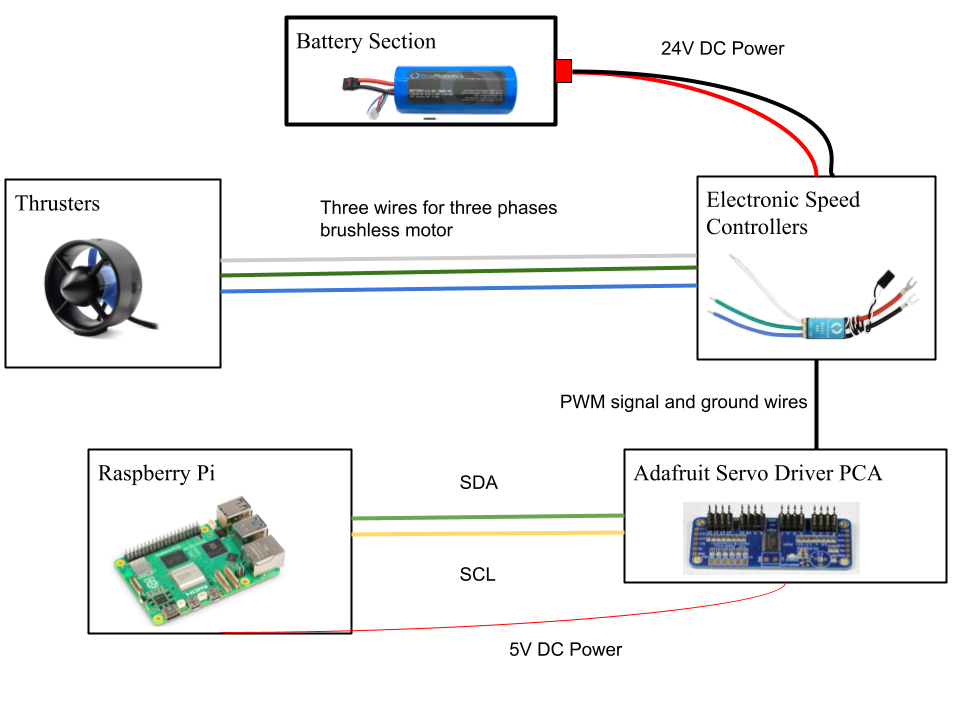}
    \caption{Control System for One Thruster}
    \label{fig:thrusterContrlSys}
\end{figure}

The role control module is to tune the speed of four thrusters. The main components in module are: four Blue Robotics T200 thrusters, four ESCs(Electronic Speed Controllers), a 16-channel servo driver and an onboard computer(Raspberry Pi). By adjusting the duty cycle of the PWM(Pulse-Width Modulation) signal, the ESC would regulate the rotation speed of the brushless motor. The control system for one thruster is shown in Fig.\ref{fig:thrusterContrlSys}. The T200 thruster powered by a 24V battery, connects to an ESC(electronic speed controller). The PWM signal is transmitted from the Servo driver. The Raspberry Pi runs thruster\_driver node, which could send commands to the servo driver by I2C communication.

In the low-level control module, the ROS node is written in Python. The Python library of the Raspberry Pi I2C interface and the Python library for Servo driver are employed to wrap of the ROS Python scripts. The control mechanism is uncomplicated. Once the motion commands received, the thrusts\_driver node would transmit the corresponding PWM signal to rotate the motor. Otherwise, the node would transmit the neutral PWM signal to keep thrusts still.

\subsection{Sensing Module}

\begin{figure}[H]
    \centering
    \includegraphics[width=0.8\linewidth]{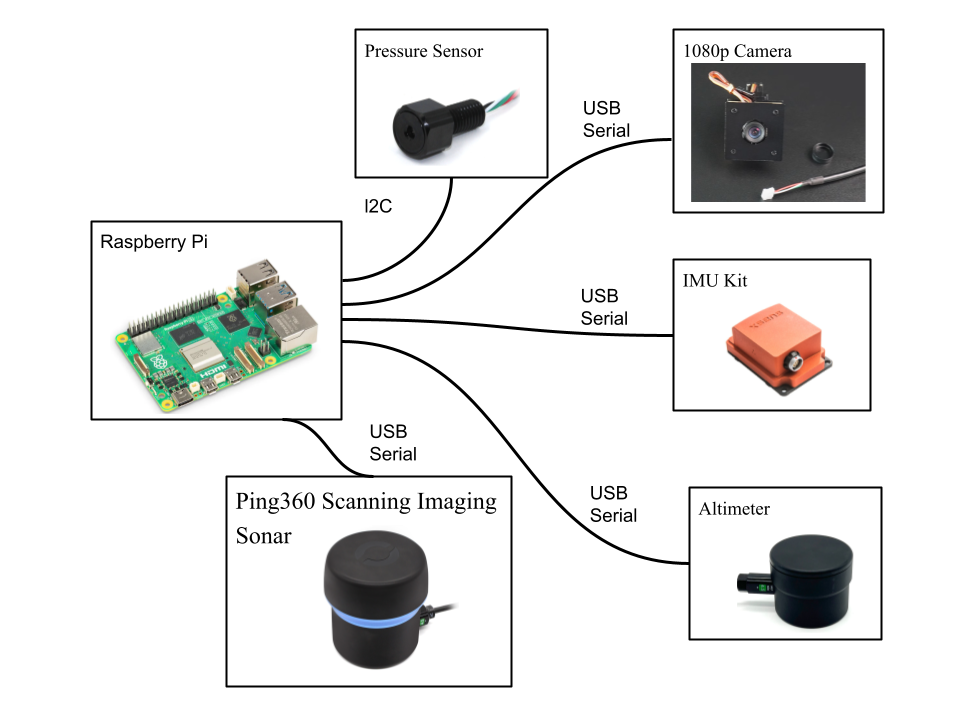}
    \caption{Layout of Sensing Module}
    \label{fig:sensor_layout}
\end{figure}

The sensing module aims to integrate the multiple sensors which are widely adopted in underwater applications. The pressure sensor is utilised to measure the depth of the vehicle. The 1080p camera provides a modest video streaming. The IMU kit measures the gyro rate, acceleration, and magnetic field. Through the strapdown algorithm, the attitudes relative to the original orientation or heading reference system (AHRS), are derived. The altimeter is a ping sonar to measure the distance between vehicle and seabed by acoustic signal. The scanning imaging sonar is used to sense the near obstacles. As depicted in Fig.\ref{fig:sensor_layout}, except the pressure sensor, the other devices connect to Raspberry Pi via the USB serial.     

Due to the big overhead of ROS on the video streaming, the GStreamer\cite{newmarch2017gstreamer} is used to transfer 1080p live streaming from IMX322 Webcam to operator terminal. As for the IMU(Xsens MTi-300), the official C++ SDK and ROS driver are provided. We have slightly modified the code, so that the complete raw data (gyroscope, accelerometer, and magnetometer) can be recorded in text form independently. For Ping2 sonar, Bar30 pressure sensor, We developed their ROS drivers  based on their office SDK. The ROS driver of Ping360 sonar has been developed by Centrale Nantes\cite{ping360sonar}.

\subsection{Navigation Module}

\begin{figure}[H]
    \centering
    \includegraphics[width=0.6\linewidth]{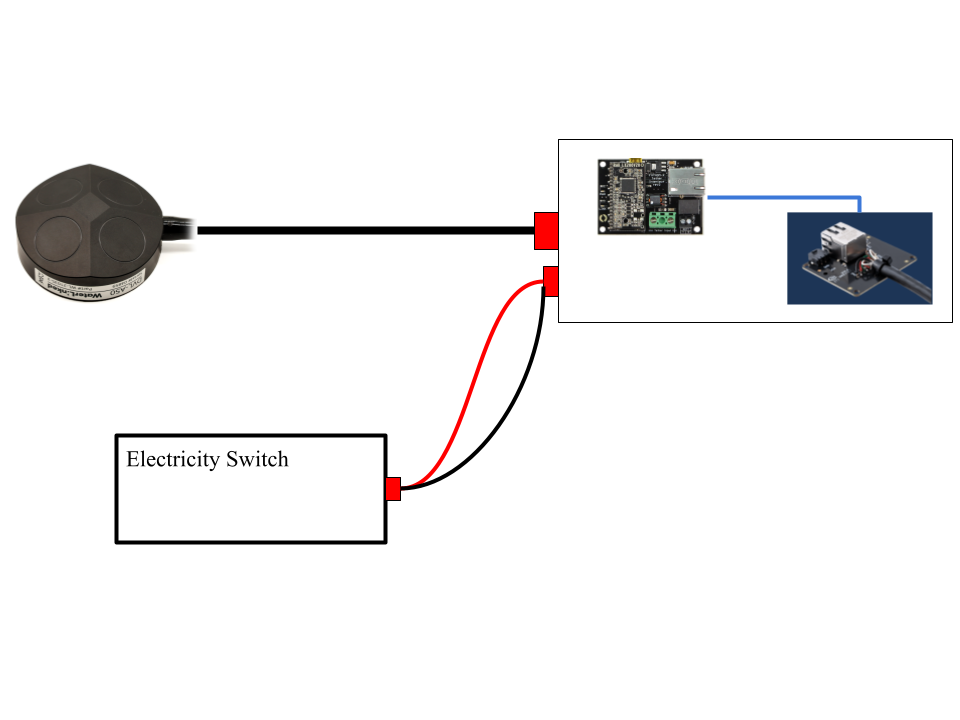}
    \caption{DVL Configuration}
    \label{fig:DVL_layout}
\end{figure}

The navigation module of YellowCube UUV contains a DVL(Doppler Velocity Log) Water Linked A50 and a USBL(Ultra Short Base Line) SeaTrac. The DVL is supposed to measure the velocities of the vehicle relative to seabed, integrating with the build-in IMU to estimate the trajectory by using the dead reckoning algorithm. The DVL configuration is depicted in Fig.\ref{fig:DVL_layout}. The SeaTrac USBL contains both an X150 USBL beacon and an X110 transponder beacon\cite{blueprint2025seatrac}. The X100 is fastened to the frame of the YellowCube UUV, and is powered by the battery. The X150 USBL beacon is fixed and connects to the operator terminal. In the case, the USBL is able to estimate the position of the UUV relative to its fixed beacon, without suffering from accumulation errors. Since the DVL and USBL both have their own communication chips and are compatible for the Ethenet protocol, the operator terminal can directly communicate with them without the need of the Raspberry Pi.

The official ROS Python driver of DVL A50 is supplied. Based on it, we modified the Python script to record all raw data and positions estimated by the built-in dead reckon algorithm. The SeaTrac USBL provides a professional software tool to record all streaming data. 

\section{Scientific Payload: Vision module}

\begin{figure}[H]
    \centering
    \includegraphics[width=0.8\linewidth]{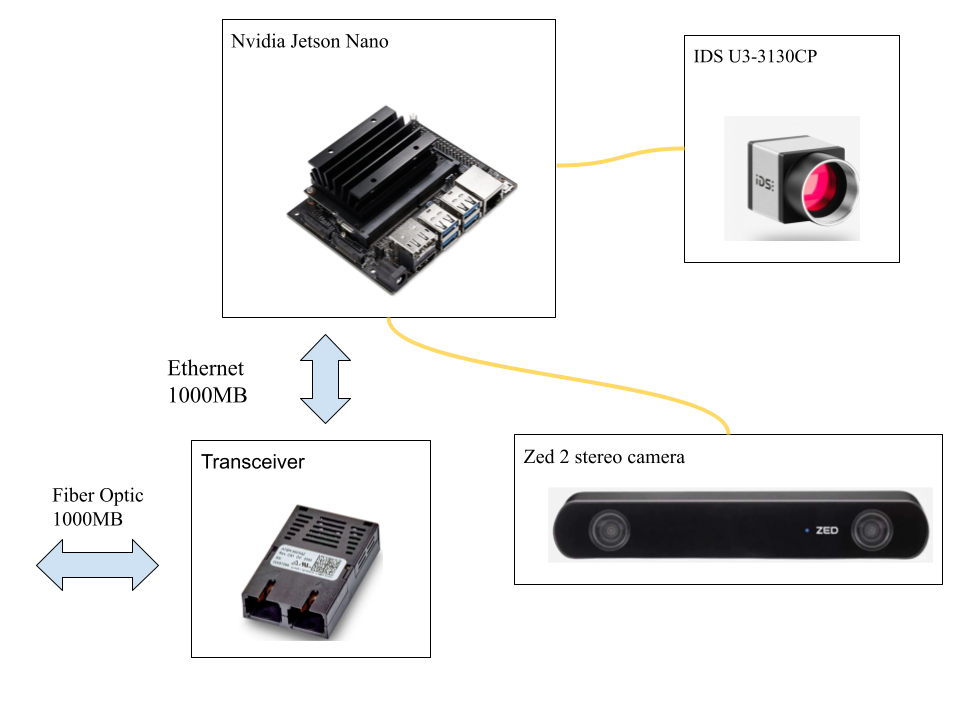}
    \caption{Computer Vision Module}
    \label{fig:scientific_payload}
\end{figure}

The development of the YellowCube UUV is intended to collect the ocean data for the study of the underwater visual navigation and perception. The vision module can be regarded as a scientific payload. The configuration of the module is shown in Fig.\ref{fig:scientific_payload}. Two main sensors are contained, one is high-frame global-shutter camera IDS U3-3130CP, another is a stereo camera Zed 2 integrating with deep learning techniques for the 3D reconstruction.    


As mentioned above, the visual information transferring would eat up the huge communication bandwidth. The vision module has its own fibre optic cable which is separated from the PLC network. A pair of single-core, single model fibre optic transceivers and fibre optical tether are used to comprise the system. The transceivers are used to transmit and receive data through the fibre optic and work as Ethernet or ATM interfaces. In practical scenarios, bandwidth of the system is able to reach over 1000Mbps. The video streaming of high-speed camera and HD stereo cameras is enabled.  

For the software part, we developed the ROS driver for IDS camera based on the official SDK, and a small programme to store images in the customised directory. As for Zed 2 stereo camera, the official ROS driver is provided, which is able to cooperate with Nvidia Jetson Nano to activate point-cloud construction function by using a deep learning method benefiting from CUDA\cite{luebke2008cuda}.   


\section{Experiments And Sea Trials}

\begin{figure}[H]
    \centering
    \includegraphics[width=0.7\linewidth]{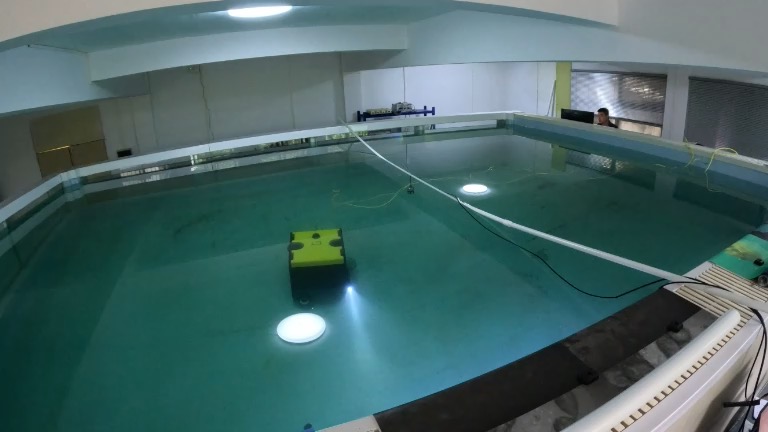}
    \caption{The Experiential Tank with the USBL beacon fixed}
    \label{fig:labExpLayout}
\end{figure}

\begin{figure}[H]
    \centering
    \includegraphics[angle=-90, width=0.3\linewidth]{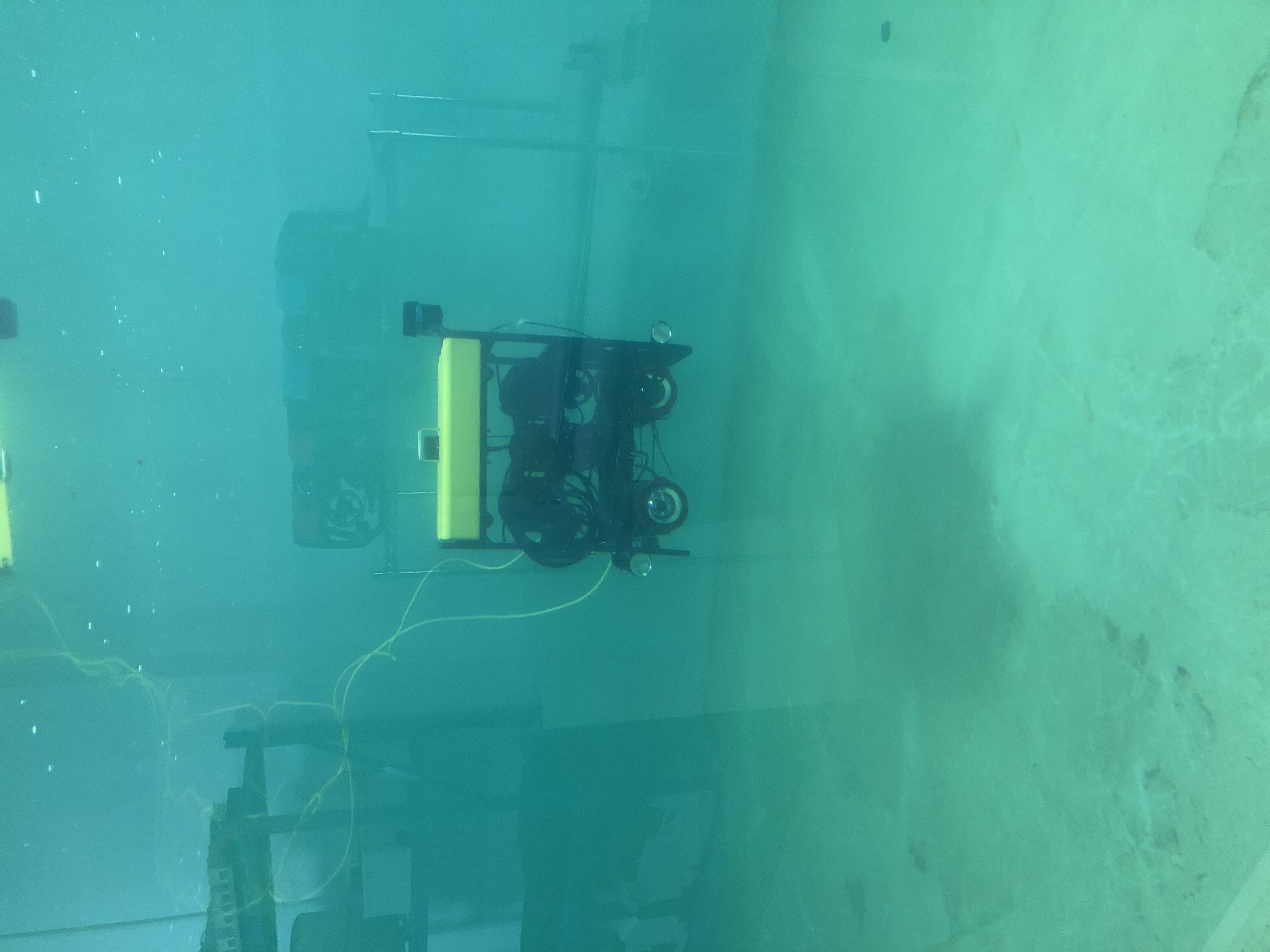}
    \includegraphics[angle=-90, width=0.3\linewidth]{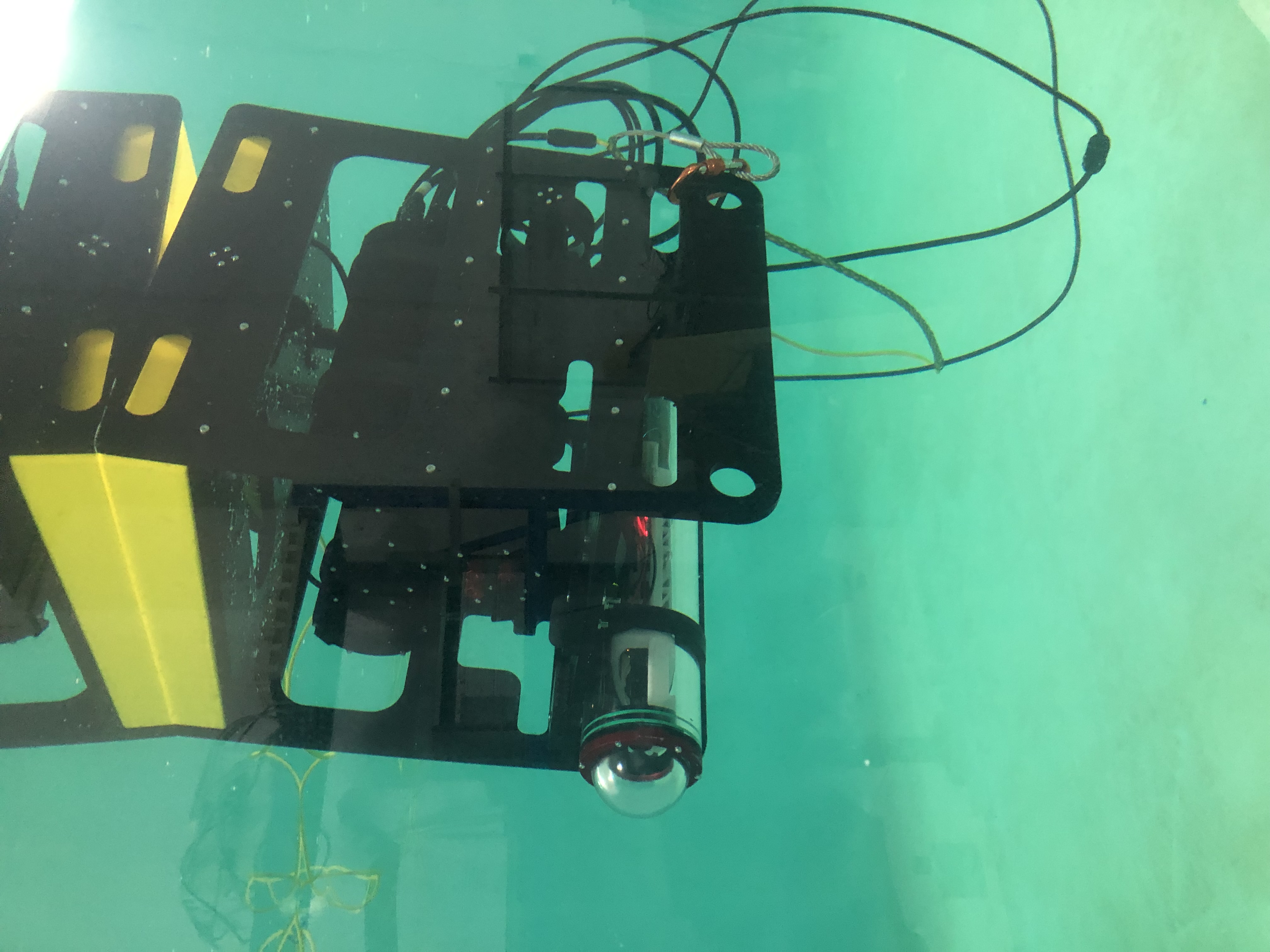}
    \caption{The Vehicle in Experiments. The right image shows the vehicle is with all modules. The left image shows the vehicle only carries the control module and the vision module}
    \label{fig:vehicleInExperiment}
\end{figure}

\begin{figure}[H]
    \centering
    \includegraphics[width=0.4\linewidth]{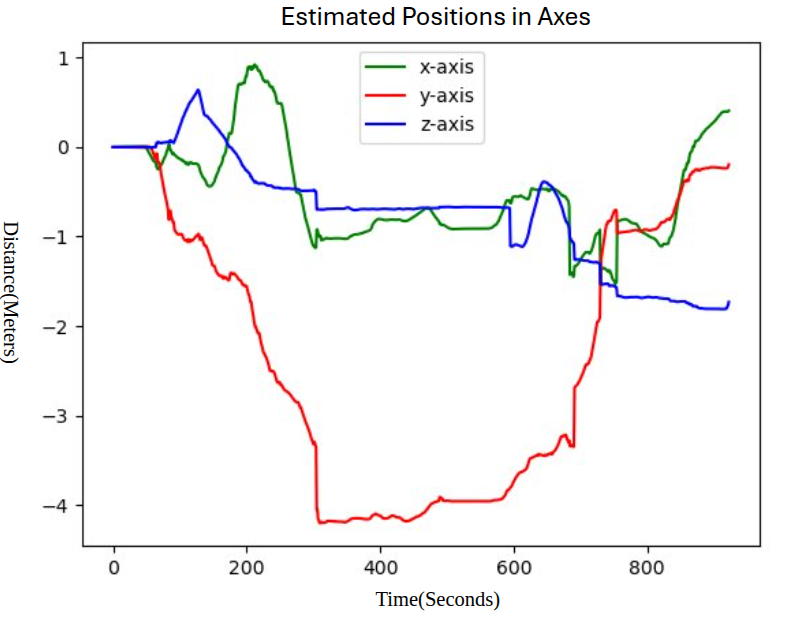}
    \includegraphics[width=0.4\linewidth]{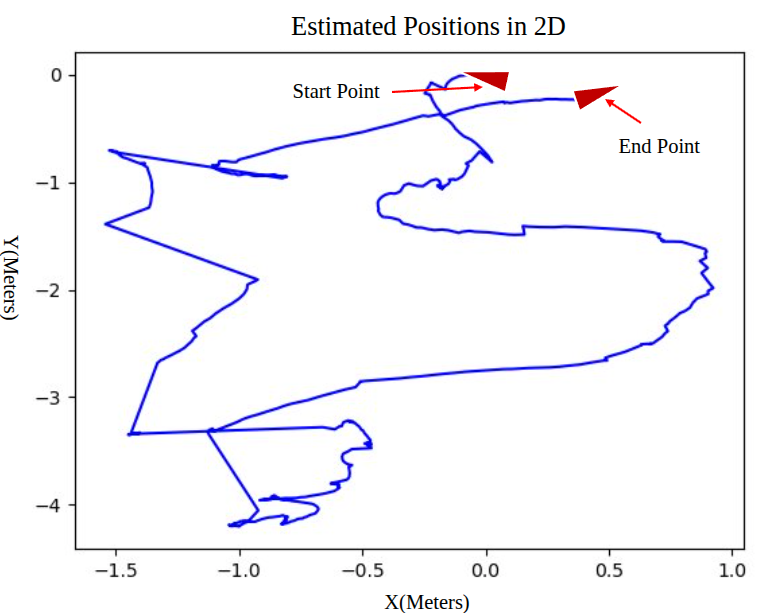}
    \caption{Position Estimated by Navigation Modules. Right plot shows the variation of the positions in three axes. Left plot shows the estimated trajectories in the x-y plane. }
    \label{fig:NavModulesPlot}
\end{figure}

The functions of modules and the P2P ROS-based communication network of the YellowCube UUV have been verified by experiments in the laboratory and sea trials. The experiments were conducted in the Key Laboratory of Intelligent Equipment for South China Sea Marine Ranching in Guangdong Ocean University. The UUV was operated in a steel-glass water tank, where the SeaTrac USBL beacon is located at the centre of the tank using a long white rob, as presented in Fig.\ref{fig:labExpLayout} and Fig.\ref{fig:vehicleInExperiment}. The vehicle is controlled by an operator via the ROS network.  All the modules worked smoothly. The trajectories estimated by the navigation module is depicted in Fig.\ref{fig:NavModulesPlot}. 

The sea trial was conducted in Zhanjiang Bay, China. During the trial, the communication was smooth. Each sensor worked normally. The thrusters could be operated to manipulate the UUV against the tide. However, the communication lag occasionally occurred in the PLC network, especially when the thrusters were in full power. Since huge energy consumed by thrusters, communication lag may be caused by the insufficient voltage. Similar problems do not occur in fibre optical communication of the vision module.

\begin{figure}[H]
    \centering
    \includegraphics[width=0.45\linewidth]{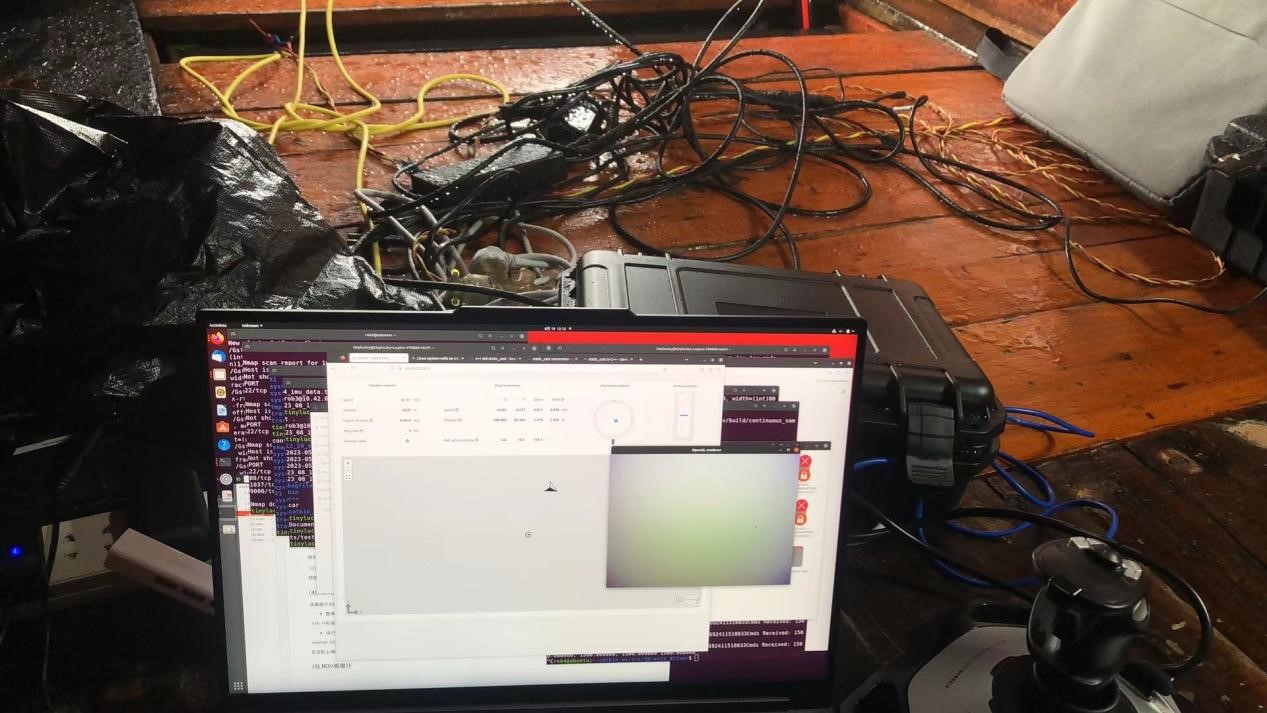}
    \includegraphics[width=0.45\linewidth]{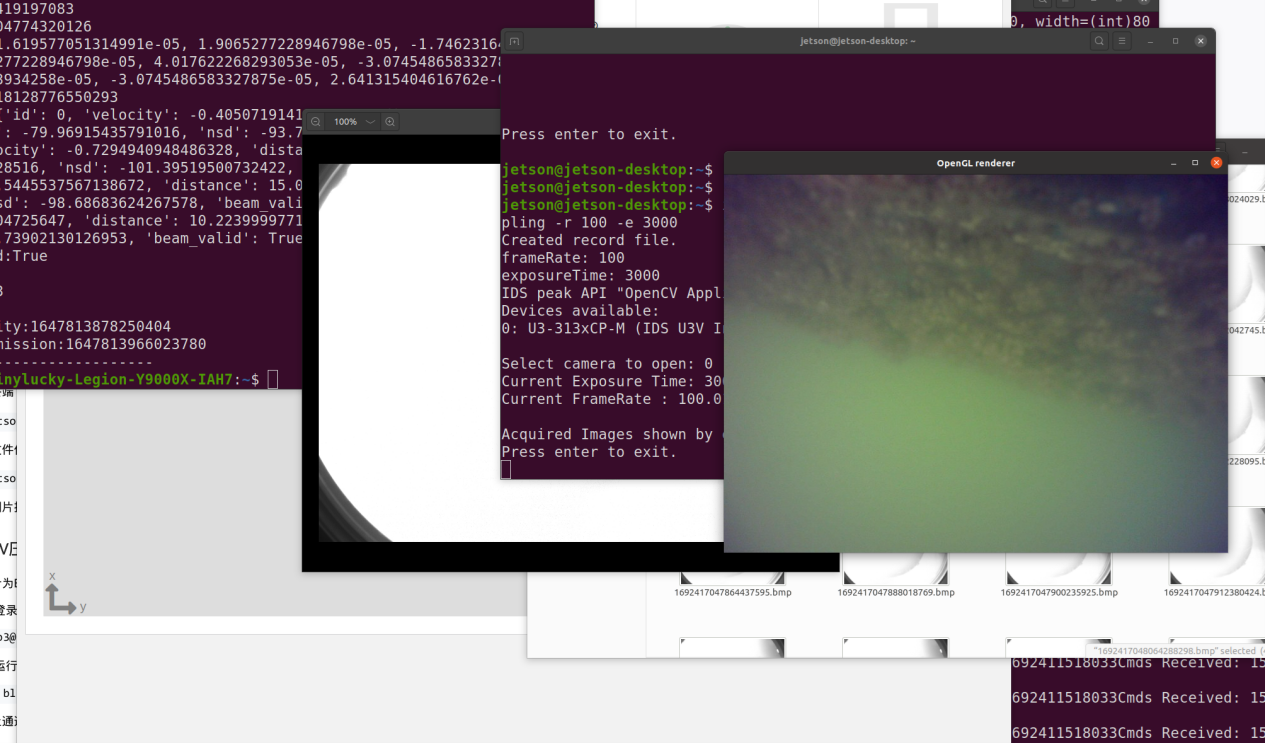}
    \caption{The Control Panel of YellowCube in Sea Trial}
    \label{fig:fieldTrial}
\end{figure}


Due to severe tide and strong current, the UUV did not dive deeply. The water in Zhanjiang Bay is relatively turbid, and the visibility of the camera in the water is extremely low. Only when it is close to the target surface can valid images be obtained. Furthermore, the colours in the images were shifting to green and yellow as shown in Fig.\ref{fig:fieldTrial}.

\section{Conclusion}

In the paper, the design and implementation of a modular and decentral UUV is presented. Compared to other similar works, a P2P communication mechanism is built up among the modules. Such configuration makes the YellowCube UUV be with more advances in updating and maintenance. The experiments and sea trials were conducted, where the UUV preformed well even under the disturbance of the strong tide. However, there are some improvements to be implemented in the future. The communication methods will be upgraded from the PLC to the fibre optics. As the more and more sensors are supported by the ROS2, the gradually switching from the ROS1 to ROS2 is going to start as well.

\section*{Acknowledgment}

The authors would like to thank Farsoon Technologies Co Ltd for providing 3D printing services.



\bibliographystyle{IEEEtran}
\bibliography{bibtex/bib/paper}

\begin{thebibliography}{10}
\providecommand{\url}[1]{#1}
\csname url@samestyle\endcsname
\providecommand{\newblock}{\relax}
\providecommand{\bibinfo}[2]{#2}
\providecommand{\BIBentrySTDinterwordspacing}{\spaceskip=0pt\relax}
\providecommand{\BIBentryALTinterwordstretchfactor}{4}
\providecommand{\BIBentryALTinterwordspacing}{\spaceskip=\fontdimen2\font plus
\BIBentryALTinterwordstretchfactor\fontdimen3\font minus \fontdimen4\font\relax}
\providecommand{\BIBforeignlanguage}[2]{{%
\expandafter\ifx\csname l@#1\endcsname\relax
\typeout{** WARNING: IEEEtran.bst: No hyphenation pattern has been}%
\typeout{** loaded for the language `#1'. Using the pattern for}%
\typeout{** the default language instead.}%
\else
\language=\csname l@#1\endcsname
\fi
#2}}
\providecommand{\BIBdecl}{\relax}
\BIBdecl

\bibitem{sangekar2008hardware}
M.~Sangekar, M.~Chitre, and T.~B. Koay, ``Hardware architecture for a modular autonomous underwater vehicle starfish,'' in \emph{OCEANS 2008}.\hskip 1em plus 0.5em minus 0.4em\relax IEEE, 2008, pp. 1--8.

\bibitem{mangayarkarasi2024modular}
T.~Mangayarkarasi, R.~Harshavardhan, R.~Sujith, and K.~Sricharan, ``A modular design approach for underwater rov: Trident,'' in \emph{2024 International Conference on Power, Energy, Control and Transmission Systems (ICPECTS)}.\hskip 1em plus 0.5em minus 0.4em\relax IEEE, 2024, pp. 1--4.

\bibitem{tolstonogov2020modular}
A.~Y. Tolstonogov, I.~A. Chemezov, A.~Y. Kolomeitsev, and V.~A. Storozhenko, ``The modular approach for underwater vehicle design,'' in \emph{Global Oceans 2020: Singapore--US Gulf Coast}.\hskip 1em plus 0.5em minus 0.4em\relax IEEE, 2020, pp. 1--7.

\bibitem{odetti2024minion}
A.~Odetti, M.~Caccia, and G.~Bruzzone, ``Minion: Modular and independent navigational intelligent orientable nozzle-thruster,'' in \emph{Conference Proceedings of iSCSS}, vol. 2024, 2024.

\bibitem{ramey2018modular}
C.~Ramey, M.~Meister, A.~Spears, J.~Lutz, D.~Dichek, B.~Hurwitz, J.~Lawrence, J.~Lawrence, M.~Philleo, and B.~E. Schmidt, ``Modular controls and instrumentation software for icefin rov,'' in \emph{OCEANS 2018 MTS/IEEE Charleston}.\hskip 1em plus 0.5em minus 0.4em\relax IEEE, 2018, pp. 1--4.

\bibitem{meinecke2011hybrid}
G.~Meinecke, V.~Ratmeyer, and J.~Renken, ``Hybrid-rov-development of a new underwater vehicle for high-risk areas,'' in \emph{OCEANS'11 MTS/IEEE KONA}.\hskip 1em plus 0.5em minus 0.4em\relax IEEE, 2011, pp. 1--6.

\bibitem{newman2008moos}
P.~Newman, ``Moos-mission orientated operating suite,'' Department of Engineering Science, University of Oxford, Tech. Rep., 2008.

\bibitem{macenski2022robot}
S.~Macenski, T.~Foote, B.~Gerkey, C.~Lalancette, and W.~Woodall, ``Robot operating system 2: Design, architecture, and uses in the wild,'' \emph{Science robotics}, vol.~7, no.~66, p. eabm6074, 2022.

\bibitem{quigley2009ros}
M.~Quigley, K.~Conley, B.~Gerkey, J.~Faust, T.~Foote, J.~Leibs, R.~Wheeler, A.~Y. Ng \emph{et~al.}, ``Ros: an open-source robot operating system,'' in \emph{ICRA workshop on open source software}, vol.~3, no. 3.2.\hskip 1em plus 0.5em minus 0.4em\relax Kobe, 2009, p.~5.

\bibitem{demarco2011implementation}
K.~DeMarco, M.~E. West, and T.~R. Collins, ``An implementation of ros on the yellowfin autonomous underwater vehicle (auv),'' in \emph{OCEANS'11 MTS/IEEE KONA}.\hskip 1em plus 0.5em minus 0.4em\relax IEEE, 2011, pp. 1--7.

\bibitem{turrisi2024decentralized}
R.~Turrisi and M.~Benjamin, ``Decentralized linear convoying for underactuated surface craft with partial state coupling,'' in \emph{2024 IEEE/RSJ International Conference on Intelligent Robots and Systems (IROS)}.\hskip 1em plus 0.5em minus 0.4em\relax IEEE, 2024, pp. 1161--1168.

\bibitem{gershfeld2023adaptive}
N.~Gershfeld, T.~M. Paine, and M.~R. Benjamin, ``Adaptive and collaborative bathymetric channel-finding approach for multiple autonomous marine vehicles,'' \emph{IEEE Robotics and Automation Letters}, vol.~8, no.~7, pp. 4028--4035, 2023.

\bibitem{paine2024model}
T.~M. Paine and M.~R. Benjamin, ``A model for multi-agent autonomy that uses opinion dynamics and multi-objective behavior optimization,'' in \emph{2024 IEEE International Conference on Robotics and Automation (ICRA)}.\hskip 1em plus 0.5em minus 0.4em\relax IEEE, 2024, pp. 8305--8311.

\bibitem{naglak2018backseat}
J.~E. Naglak, B.~R. Page, and N.~Mahmoudian, ``Backseat control of sandshark auv using ros on raspberrypi,'' in \emph{OCEANS 2018 MTS/IEEE Charleston}.\hskip 1em plus 0.5em minus 0.4em\relax IEEE, 2018, pp. 1--5.

\bibitem{bjellossubjugator}
L.~Bjellos, L.~Bonilla, C.~Brown, A.~Fernandez, A.~Hamdan, D.~Parra, and E.~Schwartz, ``Subjugator 2024: Design and implementation of a modular, high-performance auv.''

\bibitem{aaltonen2020implementation}
T.~Aaltonen, M.~Saarivirta, T.~Kerminen, and J.~Gr{\"o}nman, ``Implementation of a low-cost autonomous underwater vehicle using open source ros components with consumer class sonar technologies,'' in \emph{2020 43rd International Convention on Information, Communication and Electronic Technology (MIPRO)}.\hskip 1em plus 0.5em minus 0.4em\relax IEEE, 2020, pp. 1189--1194.

\bibitem{christ2013rov}
R.~D. Christ and R.~L. Wernli~Sr, \emph{The ROV manual: a user guide for remotely operated vehicles}.\hskip 1em plus 0.5em minus 0.4em\relax Butterworth-Heinemann, 2013.

\bibitem{rviz360aadithya}
\BIBentryALTinterwordspacing
A.~Vijayakumar. (2023) Rviz vs gazebo in ros: The importance of 3d visualization for robotics development. [Online]. Available: \url{https://ros-learnings.hashnode.dev/rviz-vs-gazebo-in-ros-the-importance-of-3d-visualization-for-robotics-development}
\BIBentrySTDinterwordspacing

\bibitem{newmarch2017gstreamer}
J.~Newmarch, ``Gstreamer,'' in \emph{Linux Sound Programming}.\hskip 1em plus 0.5em minus 0.4em\relax Springer, 2017, pp. 211--221.

\bibitem{ping360sonar}
\BIBentryALTinterwordspacing
C.~Nantes. (2020) Ping360 sonar. [Online]. Available: \url{https://github.com/CentraleNantesRobotics/ping360_sonar}
\BIBentrySTDinterwordspacing

\bibitem{blueprint2025seatrac}
\BIBentryALTinterwordspacing
B.~Subsea. (2025) Seatrac standard features. [Online]. Available: \url{https://www.blueprintsubsea.com/seatrac/seatrac-standard}
\BIBentrySTDinterwordspacing

\bibitem{luebke2008cuda}
D.~Luebke, ``Cuda: Scalable parallel programming for high-performance scientific computing,'' in \emph{2008 5th IEEE international symposium on biomedical imaging: from nano to macro}.\hskip 1em plus 0.5em minus 0.4em\relax IEEE, 2008, pp. 836--838.

\end{thebibliography}

\end{document}